\documentclass[sigconf,natbib=false,anonymous=false]{acmart}

\AtBeginDocument{%
  \providecommand\BibTeX{{%
    \normalfont B\kern-0.5em{\scshape i\kern-0.25em b}\kern-0.8em\TeX}}}

\setcopyright{acmcopyright}
\copyrightyear{xxx}
\acmYear{xxx}
\acmDOI{XX.XX}

\acmConference[xxx]{xxx}{xxx}{xxx}

\acmPrice{xxx}
\acmISBN{xxx}
\usepackage{multirow}
\usepackage{array}
\usepackage{subcaption}
\usepackage{cite}
\usepackage{amsmath}
\usepackage{url}
\usepackage{tabularx}
\usepackage{csquotes}

\begin{document}

\title{On the Uses of Large Language Models to Interpret Ambiguous Cyberattack Descriptions}

\author{Reza Fayyazi}
\email{rf1679@rit.edu}
\authornotemark[1]
\affiliation{%
  \institution{Rochester Institute of Technology}
  \streetaddress{1 Lomb Memorial Dr}
  \city{Rochester}
  \state{NY}
  \country{USA}
  \postcode{14623}
}

\author{Shanchieh Jay Yang}
\email{jay.yang@rit.edu}
\affiliation{%
  \institution{Rochester Institute of Technology}
  \streetaddress{1 Lomb Memorial Dr}
  \city{Rochester}
  \state{NY}
  \country{USA}}

\renewcommand{\shortauthors}{Fayyazi and Yang}

\begin{abstract}
The volume, variety, and velocity of change in vulnerabilities and exploits have made incident threat analysis challenging with human expertise and experience along. Tactics, Techniques, and Procedures (TTPs) are to describe how and why attackers exploit vulnerabilities. However, a TTP description written by one security professional can be interpreted very differently by another, leading to confusion in cybersecurity operations or even business, policy, and legal decisions. Meanwhile, advancements in AI have led to the increasing use of Natural Language Processing (NLP) algorithms to assist the various tasks in cyber operations. With the rise of Large Language Models (LLMs), NLP tasks have significantly improved because of the LLM's semantic understanding and scalability. This leads us to question how well LLMs can interpret TTPs or general cyberattack descriptions to inform analysts of the intended purposes of cyberattacks. We propose to analyze and compare the direct use of LLMs (e.g., GPT-3.5) versus supervised fine-tuning (SFT) of small-scale-LLMs (e.g., BERT) to study their capabilities in predicting ATT\&CK tactics. Our results reveal that the small-scale-LLMs with SFT provide a more focused and clearer differentiation between the ATT\&CK tactics (if such differentiation exists). On the other hand, direct use of LLMs offer a broader interpretation of cyberattack techniques. When treating more general cases, despite the power of LLMs, inherent ambiguity exists and limits their predictive power. We then summarize the challenges and recommend research directions on LLMs to treat the inherent ambiguity of TTP descriptions used in various cyber operations. 

\end{abstract} 




\maketitle

\section{Introduction}
With the rise and development of the Internet, many systems around the world are susceptible to severe security threats. The evolving changes in vulnerabilities have made incident threat analysis challenging, even for experienced human experts. Before the MITRE ATT\&CK framework \cite{MITRE}, the cybersecurity community relied on different databases, such as CVEs \cite{CVE} (Common Vulnerabilities and Exposures), to detect, understand, and classify security threats. However, despite the usefulness of these reference databases in providing critical information about the vulnerabilities and weaknesses, their uses are mainly about matching vulnerabilities to existing systems. 
An important, yet labor-intensive and expertise-demanding task is to interpret and comprehend the various cyber attack techniques that are constantly updated through reference models. Hence, Common Attack Pattern Enumeration and Classification (CAPEC) \cite{CAPEC} and, more comprehensively, MITRE ATT\&CK was introduced to provide insights about the behaviors and techniques utilized by adversaries when they target a system. The ATT\&CK framework utilizes the concept of TTPs (i.e., Tactics, Techniques, and Procedures) to explicate the \textbf{how} and \textbf{why} behind attackers exploiting vulnerabilities. However, these descriptions can be challenging to interpret due to their inherent ambiguity, leading security analysts to potentially draw different conclusions from the same description. For example, consider the following description:

\noindent 
\begin{quote}
{\it
Adversaries may execute their own malicious payloads by side-loading DLLs. Similar to DLL Search Order Hijacking, side-loading involves hijacking which DLL a program loads. But rather than just planting the DLL within the search order of a program then waiting for the victim application to be invoked, adversaries may directly side-load their payloads by planting then invoking a legitimate application that executes their payload(s).
}
\end{quote}
\vspace{2pt}

\noindent According to the ATT\&CK framework, there are three tactics associated with this description, namely, `Privilege-Escalation', `Defense-Evasion', and `Persistence'. It is not obvious that this DLL side-loading technique will lead to any of the three, not to mention all three adversarial intents. In fact, some might interpret this description and map it to ``Execution'' since the adversary tries to execute its malicious payload. This raises the question: can LLMs (direct use or fine-tuning) perform such a challenging task with ambiguous unstructured data?

The application of Natural Language Processing (NLP) to cybersecurity problems has been a promising development, with several works being proposed in this area \cite{Kaliyar2021, Alhogail2021, Rahali2021, Suciu2022, Shahid2021, Das2020}. There have also been some works that process TTPs by leveraging NLP with Machine Learning (ML) and Deep Learning (DL) algorithms \cite{Sauerwein2022, Husari2017, You2022, Rani2023, Alves2022}, but not LLMs. The evolution of Large Language Models (LLMs), such as the OpenAI's GPT models \cite{OpenAI}, have facilitated significant improvements in NLP tasks \cite{Teubner2023, Tamkin2021, Zhao2023}. This can be largely attributed to their capabilities in semantic understanding and scalability. It is inevitable that LLMs will be integrated into cybersecurity operations and decision making. To inform the community with a proper use and scientific expectation of using LLMs for cybersecurity operations, we examine the effectiveness of two different uses of LLMs to interpret cyberattack descriptions and map them to ATT\&CK tactics.

Specifically, we explore the advantages and limitations of directly using the state-of-the-art LLMs, such as GPT-3.5 \cite{GPT3} and Bard \cite{Bard}, versus supervised fine-tuning (SFT) of small-scale-LLMs, such as Bidirectional Encoder Representation from Transformers (BERT) \cite{Devlin2018} and SecureBERT \cite{Aghaei2023}, with labeled attack descriptions. To the best of our knowledge, this is the first formal study to compare these approaches of using LLMs to interpret cyberattack descriptions. Note that we focus on these two `practical' approaches that we believe are reasonable for cybersecurity practitioners to adopt. Based on our experimental findings using both ATT\&CK and CAPEC descriptions, we present the benefits and limitations of using current LLMs and suggest future research directions. Our main contributions are as follows:

\begin{itemize}
    \item We present the first study, to the best of our knowledge, that compares two practical uses of LLMs: direct use of LLMs (GPT-3.5 and Bard) versus supervised fine-tuning (SFT) of small-scale-LLMs (BERT and SecureBERT), for interpreting TTP descriptions.
    \item We carefully design the LLM prompts and SFT to treat the non-trivial multi-label (multi-intent) overlap and ambiguity in TTP descriptions. 
    \item We demonstrate that SFT of small-scale-LLMs provide a more precise-yet-not-flexible differentiation of cybersecurity tactics (if such differentiation exists and needed), while the direct use of LLMs offer a broad-yet-not-precise interpretation of TTP descriptions.
    \item We discover two inherent types of ambiguity in TTP descriptions and show that despite the advanced capabilities of LLMs, challenges remain to treat these ambiguities. 
    \item We recommend ways to fine-tune LLMs with prompt adaptation and future research directions with self-supervised and reinforcement learning for LLMs -- to address the inherent ambiguity of TTP descriptions, the continuously evolving vulnerabilities, and the needs for usable security for assisting cybersecurity operations.
\end{itemize}
 
The remainder of this paper is organized as follows. Section \ref{sec:related} discusses related works in LLMs, especially those applied to cybersecurity operations. Section \ref{sec:method} presents our methodology and experimental designs. Section \ref{sec:results} shows our experimental results and findings. Based on our findings, we summarize the challenges and present recommendation for future research directions in Section \ref{sec:recomm}. Section \ref{sec:conclusion} concludes our study and findings.

\vspace*{-4pt}
\section{Related Works}
\label{sec:related}
We start by discussing the state-of-the-art on Large Language Models (LLMs), how NLP is used in different cybersecurity operations, and the works that utilize NLP to process TTP descriptions. 

\subsection{Large Language Models}

The domain of Natural Language Processing (NLP) experienced a significant shift with the rise of Large Language Models (LLMs), which resulted in an unprecedented scale of language understanding and generation. One of the groundbreaking introductions was BERT \cite{Devlin2018}, a pioneer LLM pre-trained on Wikipedia and the Book Corpus, possessing around 340 million parameters. This model became a turning point, demonstrating a high-level understanding of language nuances and context with around 340 million parameters. Moreover, RoBERTa \cite{Liu2019}, built on BERT's foundation, pushed the limits of bidirectional language understanding further by refining the training methodology and with more training parameters (around 355 million).

However, the term `large' became a relative term with the advent of even larger language models, such as OpenAI's GPT-3 \cite{GPT3} and Google's Bard \cite{Bard}, with over 175 and 137 billion parameters, respectively \cite{ParamsBardGPT}. These models, trained on extensive datasets and fueled by massive computational resources, have ushered in a new era of LLMs, capable of generating human-like text and identifying complex patterns across diverse domains. These LLMs have demonstrated not just their capacity for scale, but also their profound ability to learn from massive pools of information and apply this knowledge in diverse and nuanced ways, as described in \cite{Teubner2023, Tamkin2021, Zhao2023}.

\subsection{NLP for Cybersecurity Operations}

The field of cybersecurity has witnessed significant advancements through the application of Natural Language Processing (NLP) techniques across various domains. Several studies have focused on utilizing NLP to detect and mitigate misinformation and phishing threats. The works by Kaliyar et al. \cite{Kaliyar2021}, Oshikawa et al. \cite{Oshikawa2020}, Oliveira et al. \cite{Oliveira2021}, and Heidari et al. \cite{Heidari2021} have led to advancements in identifying fake news. Similarly, for phishing detection, research works by Alhogail et al. \cite{Alhogail2021} and AbdulNabi et al. \cite{AbdulNabi2021} have enhanced the ability to flag potential phishing threats.

Another vital application of NLP lies in the classification and assessment of cyber threats. NLP has proven effective for malware classification, as highlighted by the works of Rahali et al. \cite{Rahali2021, Rahali2023} and Yesir et al. \cite{Yesir2021}. 

In addition, research studies from Marcelli et al. \cite{Suciu2022} and Yin et al. \cite{Yin2020} have harnessed NLP techniques to assess exploitability and identify vulnerabilities. Similarly, the prediction of Common Vulnerability Scoring System (CVSS) scores has been improved, as investigated by Costa et al. \cite{Costa2022} and Shahid et al. \cite{Shahid2021}.

Lastly, NLP has played a role in enhancing intrusion detection and understanding attack patterns. Studies by Moskal and Yang [14] and Das et al. [15] have demonstrated the value of NLP in this domain, highlighting its potential to improve the detection and characterization of intrusions. These diverse applications reflect the growing significance of NLP in cybersecurity.

\subsection{NLP for TTP interpretations}

In the context of processing TTP descriptions, there have been several works proposed that leverage NLP and ML techniques.

One such tool is the Threat Report ATT\&CK Mapper (TRAM) developed by MTIRE \cite{TRAM}. TRAM uses logistic regression to map unstructured text descriptions to ATT\&CK techniques, although it requires human expertise to review the classified labels. Sauerwein and Pfohl \cite{Sauerwein2022} combined ML algorithms with NLP to extract cybersecurity-related information from TTP descriptions and classify techniques and tactics. Similarly, Husari et al. \cite{Husari2017} proposed TTPDrill, which combines NLP with Information Retrieval techniques to extract threat actions. Legoy \cite{Legoy2019} tested different text representation methods and multi-label classification models for TTP classification. They found that TF-IDF bag-of-words performed best with the Linear Support Vector Machine (Linear-SVM).

Kim et al. \cite{Kim2022} and Haque et al. \cite{Haque2023} also utilized ML models to classify TTPs and map them to ATT\&CK techniques. Haque et al. used Term Frequency-Inverse Document Frequency (TF-IDF) and cosine similarity to rank the top techniques. Rahman and Williams \cite{Rahman2022} focused on identifying the co-occurrence of techniques based on the ATT\&CK framework to assist security analysts in deriving mitigation strategies.

More recently, the following works leveraged the power of small-scale-LLMs (e.g., BERT) to classify TTPs. You et al. \cite{You2022} proposed the TIM framework, which groups sets of three continuous descriptions as candidates to classify TTPs. The authors used five techniques and one tactic of the ATT\&CK framework as their TTP categories. They used SentenceBERT \cite{reimers2019sentence} to get embeddings from descriptions, and utilized TTP element features, such as IPs and URLs to improve their classification accuracy. Alves et al. \cite{Alves2022} used different variants of BERT models with different hyperparameters to select the best model to classify TTPs to the ATT\&CK framework. Rani et al. \cite{Rani2023} proposed TTPHunter, a tool to extract TTPs from unstructured text with well-known transformer models, namely BERT and RoBERTa. The authors further put a linear classifier to map the output vector into the corresponding TTP class. Orbinato et al. \cite{Orbinato2022} provided an experimental study to map TTPs from unstructured Cyber Threat Intelligence (CTI) texts. They compared traditional ML and DL algorithms, such as SVM and LSTM, with SecureBERT, and showed that SecureBERT outperformed the traditional ML models. However, they did not provide a comparison between SecureBERT and BERT to show which performs better. 

Moreover, several studies, including \cite{TRAM}, \cite{You2022}, and \cite{Husari2017}, operate on the sentence level. This approach might overlook meaningful context within the surrounding text necessary for TTP classification. Therefore, in this work, we operate on the document-level. Many of these studies also state there are not any publicly available
datasets to perform TTP classification. Furthermore, to the best of our knowledge, no current work has utilized LLMs, such as GPT-3.5, to process and classify TTP descriptions. None of the previous works dealt with multi-label classification and the overlapping situation in TTPs. None used CAPEC-vs-ATT\&CK for evaluation. None compared the two fundamental uses of LLMs and offered insights to contrast their capabilities for TTP interpretation.

\section{Methodology \& Experimental Design}
\label{sec:method}

The aim of this work is to compare a direct use of LLMs and supervised fine-tuning (SFT) of small-scale-LLMs with labeled ATT\&CK TTP descriptions to see how well they can interpret cyberattack descriptions and predict one or multiple tactic(s). This is a challenging process due to multiple reasons: 1) there are no publicly available datasets that map cybersecurity descriptions into ATT\&CK tactic(s), 2) cyberattack descriptions can map to multiple ATT\&CK tactics that small-scale-LLMs need to consider, and 3) a well-structured prompt is needed for LLMs to generate the desired outputs. Therefore, in this section, we talk about our methodology and design of experiments to tackle these challenges. 

\begin{figure*}[t]
\centering
\includegraphics[scale=0.35]{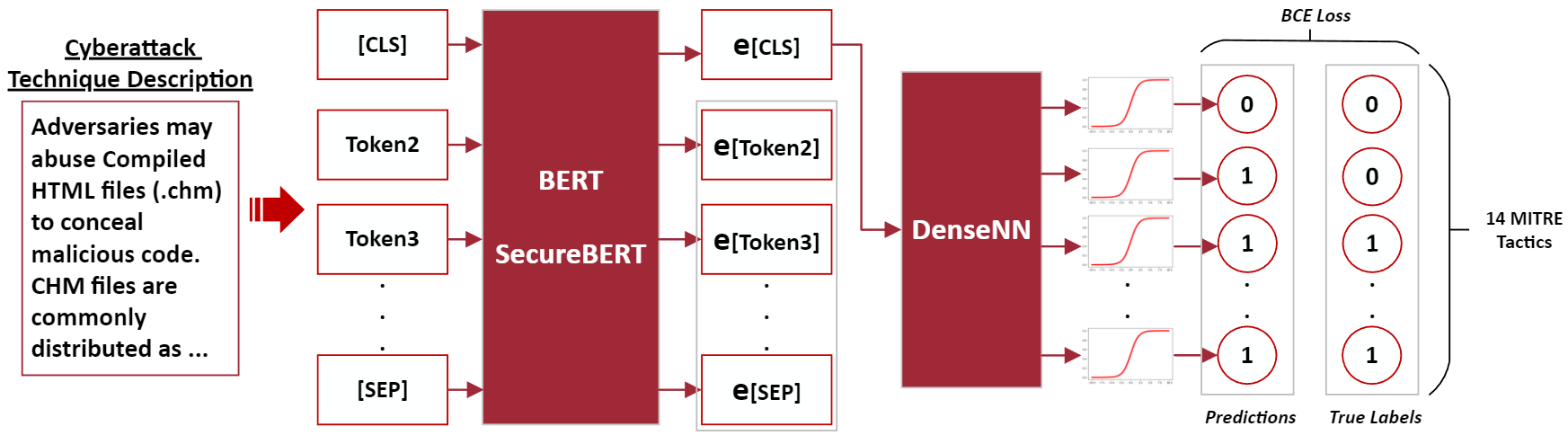}
\caption{The overall process of supervised multi-label fine-tuning of small-scale-LLMs with ATT\&CK descriptions.}
\label{fig:supervisedtraining}
\end{figure*}

\subsection{Datasets}

To conduct our experiments, we gathered data from both ATT\&CK \cite{MITRE} and CAPEC \cite{CAPEC} frameworks. We chose ATT\&CK for training due to its broad adoption in industry SIEM tools and associated hundreds of descriptions. Interpreting the 14 tactics gives a more tractable problem than resolving the 196 techniques or 411 sub-techniques while offering more precision than the common 3-7 killchain stages. In the MITRE ATT\&CK framework, we curated the descriptions of enterprise tactics, techniques, and sub-techniques along with the mapping to their corresponding tactics. Some descriptions have 2,3, or up to 4 corresponding tactics. In total, we obtained 618 descriptions from the MITRE ATT\&CK framework. Note that the use of 618 labeled ATT\&CK descriptions for supervised fine-tuning of small-scale-LLMs reflects the real-world scenario where only limited labeled data would exist. We will share the curated data along with our source code as artifacts for the research community. 

We develop a set of pair-wise overlap diagrams in Figures \ref{fig:originalOverlapMITRE} (ATT\&CK) and \ref{fig:originalOverlapCAPEC} (CAPEC) to highlight the inherent tactic overlaps (ambiguity) exhibited by the TTP descriptions. 
These diagrams with actual overlaps are placed later in the paper to make it easier to compare to the LLM prediction results. The diagonal line in these overlap diagrams show the number of descriptions that have one and only one tactics, while the symmetrical upper and lower triangles show the number of descriptions that have at least the corresponding pair of tactics. The last two columns gives the `total number of descriptions that contain the specific tactic' and `the percentage of descriptions that has at least one other tactic'. 

Upon closer observation of the actual ATT\&CK overlap, it is evident that a significant number of TTP descriptions incorporate `Persistence', `Privilege-Escalation', and/or `Defense-Evasion' tactics. In fact, `Privilege-Escalation' appears as the sole tactic in only 3 out of 97 descriptions, implying that the remaining descriptions incorporate at least one additional relevant tactic.

In the CAPEC framework, we found descriptions that were deemed relevant to the MITRE ATT\&CK taxonomy by human experts. The use of CAPEC serves to provide a critical evaluation to explore a scenario where the model must adapt to new/unseen TTP descriptions (i.e., same task with a different test-set). Out of the total 593 CAPEC descriptions, there are 177 ones that reference to techniques or sub-techniques in the ATT\&CK framework. Therefore, we curated the descriptions of those 177 CAPECs along with the set of the mapping to their corresponding ATT\&CK tactics. Note that CAPEC precedes ATT\&CK and is not meant to mapped to it. The relevance is at best an indirect reflection of human's interpretation of how the specific CAPEC descriptions were interpreted by human experts, which we expect some inconsistencies. 

Figure \ref{fig:originalOverlapCAPEC} shows the pair-wise overlap between the tactics for the CAPEC descriptions. There seems to be even more overlaps in CAPEC descriptions -- matching our expectation where human interpretation of the (ambiguous) TTP type of descriptions may differ from one to another. It is worth noting that we collected the ATT\&CK and CAPEC data in May 2023. 

Lastly, to evaluate the LLMs' performance on generic and non-cybersecurity cases, we gathered 100 IMDB reviews (which are very different from a cyberattack description) \cite{IMDB} to confirm whether the LLMs can predict `NONE' for these. 

\subsection{Supervised Fine-Tuning of Small-scale LLMs}

To do supervised fine-tuning (SFT) with small-scale-LLMs, we make use of two well-known pre-trained transformer models, namely BERT-base \cite{Devlin2018}, and SecureBERT \cite{Aghaei2023} (which is based on the RoBERTa-base \cite{Liu2019} model). We chose these models as prior works showed that these models tend to perform significantly better than traditional ML and DL algorithms in processing TTP descriptions \cite{Rani2023, Alves2022, Orbinato2022}. Note that we used the base version of these pre-trained small-scale models, as they offer a better trade-off between performance, speed, and computational power. This is a practical option for cybersecurity professionals rather than retraining the large scale LLMs. We also recognize that in reality, as it is the case for this research study, it is unlikely a very large amount of TTP descriptions with tactics will be available. Thus we focus on a targeted fine-tuning approach that treats multi-label data.

Both of these models have 12 transformer layers and 12 attention heads with hidden size of 768, and the feed-forward size of 3072. BERT-base has around 110 million parameters, while SecureBERT has around 125 million parameters. Moreover, the SecureBERT model has already being fine-tuned with a large corpus of cybersecurity texts (CVE descriptions). Therefore, to test the capability of both of these models in capturing the essential information within cybersecurity descriptions and classifying them into their corresponding ATT\&CK tactics, we fine-tune them with the curated MITRE ATT\&CK dataset.  It is important to note that due to BERT's 512-token input limit, we truncated the ends of the descriptions that exceeded this limit. This condition was found in six instances within the ATT\&CK descriptions.

Figure \ref{fig:supervisedtraining} shows the overall process of fine-tuning both BERT and SecureBERT models for classification. First, we tokenize the ATT\&CK descriptions into the desired format for both BERT and SecureBERT, and we load both of these pre-trained models. Second, we add a classification layer (DenseNN) to the models with the size of 14 (reflecting the number of ATT\&CK tactics). To treat this multi-label classification problem, we apply the Sigmoid activation function to get the probability of each tactic independently for a given description. This approach allows an effective fine-tuning even with a relatively small amount of multi-labeled unstructured TTP descriptions. 

Table \ref{tab:hyperparameters} shows the hyperparameters we selected to fine-tune the models. We employed 5-fold cross validation, carrying out 25 epochs for each fold, under a fixed random state of 42 to ensure consistent partitioning of the dataset. AdamW \cite{zhuang2022understanding},
a refined version of the Adam optimizer, was selected for the optimization process. Binary Cross-entropy served as the loss function, due to the binary nature of our classification task. To regulate the learning rate during training, we utilized a scheduler with zero warm-up steps and an initial rate of 2e-5. We selected this value of learning rate as Sun et al. \cite{sun2019chinese} found that works well for most tasks. To ensure that our training steps fit into GPU memory, we employed a batch size of 8. For training the models, we set the random seed to 519 to ensure reproducibility. During this process, we fine-tuned the pre-trained models from scratch. Therefore, the weights of the pre-trained models are adjusted slightly, and the weights of the classification layers are learned from scratch. Then, we evaluate the performance of each fold with its validation set.
The output vector is a 14-dimensional binary vector, with zeros and ones indicating the values for each ATT\&CK tactic. In the output vector, the labels with a probability value greater than 0.5 are translated as `1', while those with probability values less than or equal to 0.5 are represented as `0'. This process allows for a clear and concise prediction of tactics corresponding to each description. 

\begin{table}[htbp]
\centering
\small
\caption{The key hyperparameters for the supervised fine-tuning of BERT and SecureBERT.}
\label{tab:hyperparameters}
\begin{tabular}{c|c}
\textbf{Hyperparameter} & \textbf{Value} \\
\hline
Optimizer & AdamW \\
Learning rate & 2e-5 \\
Loss Func. & BCE \\
Activation Func. & Sigmoid \\ 
Batch Size & 8 \\
Epoch & 25 \\

\end{tabular}
\end{table}

\subsection{Prompt Engineering}

Prompt Engineering is a new concept in NLP, specifically in the context of LLMs, such as GPT-3.5. A prompt is an input that we feed into the model. It can be a question, a statement, or an incomplete sentence that the model is expected to either respond to it or complete. Prompt engineering is the design of effective and to-the-point inputs to guide the model in generating the desired output. The form and the structure of the prompts can affect the generated text significantly. A prompt usually consists of three components: 
\begin{itemize}
    \item \textbf{Task}: Provide a detailed instruction of what the model is supposed to do:
    \item \textbf{Example}: It has shown that providing examples in the prompt can help the model to adapt and generate a more desired output\cite{Fine-tuning}. This is called `few-shot learning'.
    \item \textbf{Data}: The data to give to the model as input.  
\end{itemize}

In the context of asking the LLMs, in particular GPT-3.5 and Bard, to predict the ATT\&CK tactics given a cyberattack technique description, we engineered our prompt after several iterations to get to the reported results. We note that there could be even a better engineered prompt that generate better results. For our experiments, we used the API of OpenAI's GPT-3.5-turbo model \cite{OpenAI} and the Bard-API GitHub repository\cite{Bard-API}. Moreover, in OpenAI models, their inherent design are non-deterministic. This means that identical inputs can result in different outputs. Therefore, to manage this aspect and have consistency, we have set the temperature parameter to "0". While this adjustment makes the outputs mostly deterministic, a minimal degree of variability could still persist \cite{OpenAI}. In Bard-API \cite{Bard-API}, however, there were no temperature argument to change. Therefore, we state that re-generating the prompt in Bard may result in different outputs. Below is the final engineered prompt that we used and gave it to the LLM models:

\noindent 
\begin{quote}
{\it
You will be given a number of descriptions delimited by triple backticks and you have to predict which MITRE ATT\&CK tactic(s) each description relates to. If a description does not relate to any of the MTIRE ATT\&CK tactics, simply predict it as [NONE].
\vspace{3pt}

There are 14 MITRE ATT\&CK Enterprise tactics in total and their names are:
\vspace{2pt}

- COLLECTION

- COMMAND\_AND\_CONTROL

- CREDENTIAL\_ACCESS

- DEFENSE\_EVASION

- DISCOVERY

- EXECUTION

- EXFILTRATION

- IMPACT

- INITIAL\_ACCESS

- LATERAL\_MOVEMENT

- PERSISTENCE

- PRIVILEGE\_ESCALATION

- RECONNAISSANCE

- RESOURCE\_DEVELOPMENT
\vspace{2pt}

Here are some examples of how you should do it:
\vspace{2pt}

1. Adversaries may circumvent mechanisms designed to control elevate privileges to gain higher-level permissions. Most modern systems contain native elevation control mechanisms that ... .

Tactic(s): EXFILTRATION 
\vspace{2pt}

2. Adversaries can use stolen session cookies to authenticate to web applications and services. This technique bypasses ... .

Tactic(s): LATERAL\_MOVEMENT, DEFENSE\_EVASION

\vspace{2pt}

Write the output in the following format:

Tactic(s): ...
}
\end{quote}
\vspace{2pt}

Based on our observations, the engineered prompt worked well in outputting the tactics in the desired format for the OpenAI's GPT-3.5 model. However, in Bard, the model was not solely predicting the ATT\&CK tactics, and it was generating additional information for each description. Therefore, in the prompt for Bard, we added the following sentence to reduce getting additional information: "Do NOT add any other information in your answer and ONLY print the tactics' names.". This caused the Bard model to pay more attention in generating the tactics' names, however, it was still generating information other than the tactics' names for most of the descriptions. Moreover, in the prompt, we asked the model to generate NONE as the label when it sees sentence that is not describing any of the MTIRE tactics. This is to deal with generic and non cybersecurity-related sentences (e.g., IMDB reviews).

It is worth noting that we conducted our experiments on a system equipped with an Intel Xeon W-2145 CPU (8 cores, 3.7GHz), an NVIDIA GeForce RTX 3060 GPU with 12 GB memory, and 32 GB of RAM. We used Python 3.10.8 for all scripts. The machine learning models were implemented using PyTorch 1.13.0 with CUDA 11.7.

\section{Results \& Discussion}
\label{sec:results}

We now present how LLMs and SFT of small-scale-LLMs perform for ATT\&CK and CAPEC descriptions. We first evaluate which small-scale-LLM models and which LLM models perform better in Sections 4.1 and 4.2, respectively. Then, in Sections 4.3 and 4.4. we discuss the capabilities and limitations of the selected small-scale-LLM and LLM models on MTIRE and CAPEC descriptions, respectively. Finally, we show specific descriptions that were not predicted well by the LLM models to illustrate the limitations in Section 4.5.

To evaluate the performance of the small-scale-LLM and LLM models, we use the Micro-F1 score and the Accuracy metrics. The Accuracy for both models indicate that the model should predict all labels correctly (exact match), and Micro-F1 score is the aggregation of the contributions of all classes to compute the metric. We also use Precision, Recall, and F1 score to show the performance for each individual ATT\&CK tactic.


\subsection{Evaluation of SFT of Small-scale-LLMs}
We test the small-scale-LLMs (BERT-SFT and SecureBERT-SFT) using the MITRE ATT\&CK dataset through k-fold cross validation. Table \ref{table:kfoldresults} shows the average results of the 5-fold cross validation for both BERT-SFT and SecureBERT-SFT. Note that the average is over all 14 tactics when cross-validated over the 618 ATT\&CK descriptions. Some tactics see better results than others, which we will discuss later. The overall accuracy and F1-scores (at the 0.8 and 0.9 level) exhibited by these models indicate reasonably capability to interpret and predict ATT\&CK tactics by both small-scale-LLM models with SecureBERT-SFT showing small advantage over BERT-SFT. This is expected since SecureBERT was exposed to cybersecurity texts before the classifier training with ATT\&CK descriptions. In comparing to the reported performance of similar works (which perform at 0.6 to 0.7 level), our small-scale-LLMs with SFT show promising performance, considering the complexity of the multi-label classification problem involving the 14 tactics.

\begin{table}[htbp]
\small
\begin{center}
\caption{The average results of k-fold cross-validation of the SFT of small-scale-LLM models for the ATT\&CK descriptions.}
\begin{tabular}{c | c | c} 
 Model & BERT-SFT & SecureBERT-SFT \\ 
\hline
 Avg. Train Loss & 0.04 & 0.04 \\
 Avg. Test Loss & \textbf{0.06} & 0.07 \\
 Avg. Accuracy & 0.76 & \textbf{0.85} \\
 Avg. Micro F1 & 0.87 & \textbf{0.92} \\
\end{tabular}
\label{table:kfoldresults}
\end{center}
\end{table}

We further tracked the models during the 5-fold validation process, and selected the best-performing model with the highest micro avg. F1-score. This best-performing model will be employed for subsequent evaluations. To illustrate the notable performance of the SecureBERT-SFT, we employed the test set from the same fold where it performed the best. We measured its F1-score on this test set (descriptions not seen through the training process) and compared these values with those of BERT, as displayed in Table \ref{table:kfoldmitrepertactic}. It can seen clearly that the fine-tuned SecureBERT-SFT in this particular fold of cross validation perform better. We will thus use SecureBERT-SFT to report for the comparative study with the direct use of LLMs given the space limitation. Note that SecureBERT-SFT performs better than BERT-SFT in most cases (but not in all cases). Generally speaking, our subsequent discussions using SecureBERT-SFT applies to BERT-SFT as well. 

\begin{table}[t]
\small
\begin{center}
\caption{Per-tactic performance of SecureBERT-SFT's best-performing model and BERT-SFT for the unseen ATT\&CK descriptions of that specific fold.}
\begin{tabular}{@{}c !{\vrule width 1pt} c | c | c@{}} 
  & {\textbf{BERT-SFT}} & {\textbf{SecureBERT-SFT}} &\\ 
\hline
\textbf{Tactics} & \textbf{F1 Score} & \textbf{F1 Score} & \textbf{Support} \\ 
\hline
 Collection & \textbf{0.88} & 0.82 & 9\\
 Command and Control & 0.92 & \textbf{1.00} & 7  \\
 Credential Access & 0.91 & \textbf{0.96} & 12 \\
 Defense Evasion & 0.93 & \textbf{0.98} & 27 \\
 Discovery & 0.82 & 0.82 & 10\\
 Execution & 0.86 & \textbf{1.00} & 4\\
 Exfiltration & 0.75 & 0.75 & 5\\
 Impact & 0.36 & \textbf{0.88} & 9\\
 Initial Access & 0.00 & \textbf{0.80} & 3\\
 Lateral Movement & 0.00 & 0.00 & 7\\
 Persistence & \textbf{0.97} & 0.94 & 17\\
 Privilege Escalation & 0.97 & \textbf{1.00} & 17\\
 Reconnaissance & 1.00 & 1.00 & 11\\
 Resource Development & 0.95 & 0.95 & 10 \\
 \hline
 Micro Avg. & 0.87 & \textbf{0.92} & 148 \\
\end{tabular}
\label{table:kfoldmitrepertactic}
\end{center}
\end{table}

\subsection{Direct Use of LLMs: GPT vs. Bard}

To evaluate the performance of Large Language Models (LLMs), we use the engineered prompts to ask GPT-3.5 and Bard to predict the ATT\&CK tactics for all 618 ATT\&CK descriptions. GPT-3.5 was able to predict almost all the descriptions with the specified format, except one instance where GPT-3.5 predicted `Remote-Access', which was not part of the pre-defined ATT\&CK tactics.

On the other hand, Bard demonstrated a somewhat varied performance. It predicted not-ATT\&CK tactics for 45 descriptions, such as `Exploitation', `Defacement', `Impair-Defenses', `DOS', and so forth. However, for the sake of consistency and to make a comparison between the two LLMs, we chose to disregard these outlier predictions and focused on the 14 ATT\&CK tactics. Although this may not entirely be a fair comparison, it does highlight a potential limitation of the direct use of LLMs.

Table \ref{table:LLMseval} shows the performance results of both LLM models on the MITRE ATT\&CK dataset. Since GPT-3.5 performs better than Bard, and with Bard generating a lot of non-conforming responses, we elect to use GPT-3.5 for the comparative study for the remaining of this paper. Note that the scores shown in Table \ref{table:LLMseval} is at a much lower level than those achieved by the SFT of small-scale-LLMs, and we will discuss in more details next.

\vspace*{-4pt}
\begin{table}[htbp]
\small
\begin{center}
\caption{Overall performance of GPT-3.5 vs Bard for all ATT\&CK descriptions.}
\begin{tabular}{c | c | c} 
 \textbf{Model} & \textbf{GPT-3.5} & \textbf{Bard} \\ 
\hline
 \textbf{Micro F1} & \textbf{0.67} & 0.66 \\
 \textbf{Accuracy} & \textbf{0.44} & 0.31 \\
\end{tabular}
\label{table:LLMseval}
\end{center}
\end{table}

\subsection{LLM vs. SFT of small-scale-LLMs (ATT\&CK)}
Building upon the discussions in Sections 4.1 and 4.2, we selected SecureBERT-SFT and GPT-3.5 as our models of choice to evaluate on the entire ATT\&CK descriptions. It is important to note that we deployed SecureBERT-SFT's best-performing model, which had already been trained on $\sim$80\% of the dataset, and tested it on the entire MITRE ATT\&CK dataset. Previously, we showed its performance on the remaining 20\% of the data in Table \ref{table:kfoldmitrepertactic}. We adopted this approach to draw a meaningful comparison with the GPT-3.5, in which, having been trained on Internet data, shares a similar exposure advantage. The purpose is to study how a SFT of small-scale-LLMs, with less parameters but focused on a specific task, fares with an LLM. 

\begin{figure}
     \centering
     \begin{subfigure}[b]{0.4\textwidth}
         \centering
         \includegraphics[width=\textwidth]{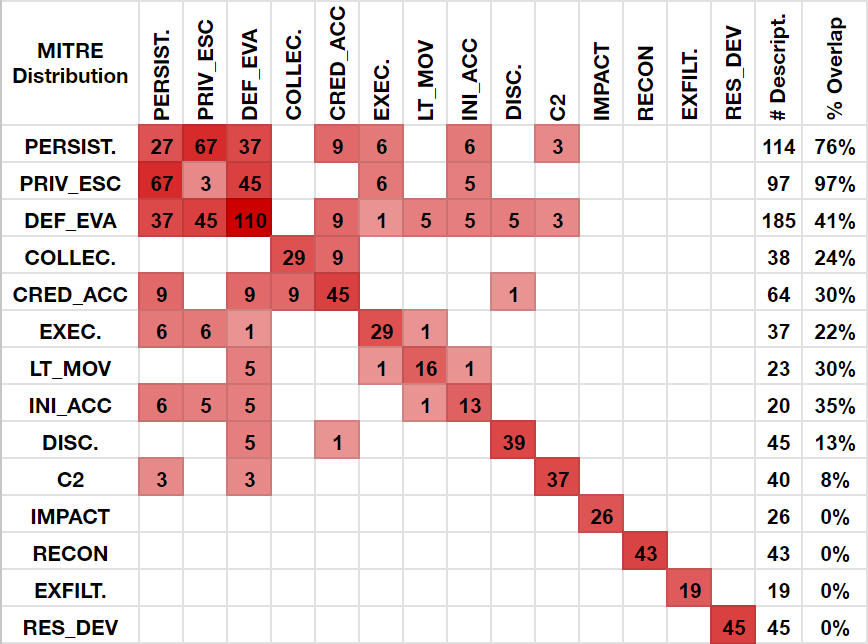}
         \caption{Actual overlaps of ATT\&CK descriptions}
         \label{fig:originalOverlapMITRE}
     \end{subfigure}
     \hfill
     \begin{subfigure}[b]{0.4\textwidth}
         \centering
         \includegraphics[width=\textwidth]{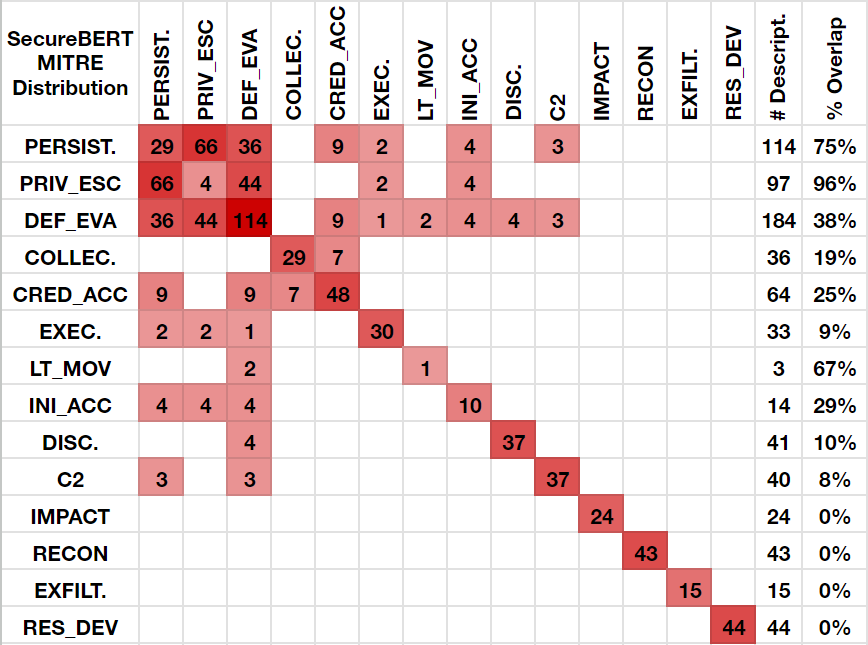}
         \caption{Tactic overlaps based on SecureBERT predictions}
         \label{fig:secureBERToverlapCAPEC}
     \end{subfigure}
     \hfill
     \begin{subfigure}[b]{0.4\textwidth}
         \centering
         \includegraphics[width=\textwidth]{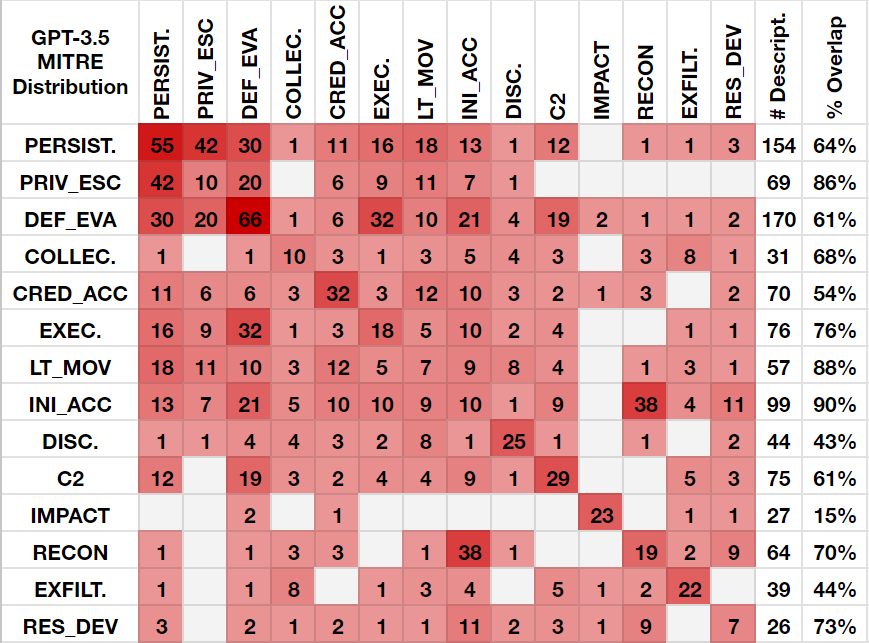}
         \caption{Tactic overlaps based on GPT-3.5 predictions}
         \label{fig:GPT3overlapMITRE}
     \end{subfigure}
        \caption{Pair-wise tactic overlap for ATT\&CK descriptions.}
        \label{fig:MITREoverlapResults}
\end{figure}

Table \ref{table:mitreoverall} summarizes the performance and show that SecureBERT-SFT, when trained with labeled data, outperforms GPT-3.5 in predicting ATT\&CK Tactics. For a deeper comparison, we laid out per-tactic precision, recall, and F1-score for SecureBERT and GPT-3.5 models in Table \ref{table:mitrepertactic}. Based on the results, we can see that SecureBERT-SFT shows a superior performance over GPT-3.5, for both tactics with high overlap and those with less or no overlap. Interestingly, SecureBERT-SFT performed poorly on descriptions with the `Lateral-Movement' tactic, for which we conjecture, by reading through the descriptions, that `Lateral-Movement' techniques significantly intertwined with other tactics even though ATT\&CK taxonomy does not suggest so. On the other hand, GPT-3.5 excelled singularly for the `Impact' tactic, matching SecureBERT-SFT's performance in this regard. This might be attributed to the more direct and unambiguous nature of descriptions for this particular tactic, in which can be supported by Figure \ref{fig:GPT3overlapMITRE}. Figure \ref{fig:originalOverlapMITRE} shows actual overlaps in which indicates that the `Impact' tactic has no overlaps with any other tactics.

\begin{table}[htbp]
\small
\begin{center}
\caption{Overall Performance of SecureBERT-SFT and GPT-3.5 for all ATT\&CK descriptions.}
\begin{tabular}{c | c | c} 
\textbf{Model} & \textbf{SecureBERT-SFT} & \textbf{GPT-3.5} \\ 
\hline
 \textbf{Micro-F1} & \textbf{0.97} & 0.67 \\
\hline
 \textbf{Accuracy} & \textbf{0.92} & 0.44 \\
\end{tabular}
\label{table:mitreoverall}
\end{center}
\end{table}

\begin{table*}[hbt]
\small
\begin{center}
\caption{Per-tactic performance of SecureBERT-SFT and GPT-3.5 for all ATT\&CK descriptions.}
\begin{tabular}{@{}c !{\vrule width 1pt} c | c | c !{\vrule width 1pt} c | c | c !{\vrule width 1pt} c@{}} 
  & \multicolumn{3}{c}{\textbf{SecureBERT-SFT}} & \multicolumn{3}{c}{\textbf{GPT-3.5}} \\ 
\hline
\textbf{Tactics} & \textbf{Precision} & \textbf{Recall} & \textbf{F1 Score} & \textbf{Precision} & \textbf{Recall} & \textbf{F1 Score} & \textbf{Support} \\ 
\hline
 Collection & 0.97 & 0.92 & \textbf{0.95} & 0.68 & 0.55 & 0.61 & 38 \\
 Command and Control & 1.00 & 1.00 & \textbf{0.95} & 0.51 & 0.97 & 0.67 & 40  \\
 Credential Access & 0.98 & 0.98 & \textbf{0.98} & 0.79 & 0.86 & 0.82 & 64 \\
 Defense Evasion & 1.00 & 0.99 & \textbf{1.00} & 0.75 & 0.69 & 0.72 & 185 \\
 Discovery & 1.00 & 0.91 & \textbf{0.95} & 0.73 & 0.71 & 0.72 & 45\\
 Execution & 1.00 & 0.89 & \textbf{0.94} & 0.39 & 0.81 & 0.53 & 37 \\
 Exfiltration & 1.00 & 0.79 & \textbf{0.88} & 0.49 & 1.00 & 0.66 & 19 \\
 Impact & 1.00 & 0.92 & 0.96 & 0.96 & 1.00 & \textbf{0.98} & 26 \\
 Initial Access & 1.00 & 0.70 & \textbf{0.82} & 0.19 & 0.95 & 0.32 & 20  \\
 Lateral Movement & 1.00 & 0.13 & 0.23 & 0.39 & 0.96 & \textbf{0.55} & 23 \\
 Persistence & 0.99 & 0.99 & \textbf{0.99} & 0.68 & 0.92 & 0.78 & 114 \\
 Privilege Escalation & 1.00 & 1.00 & \textbf{1.00} & 0.72 & 0.52 & 0.60 & 97  \\
 Reconnaissance & 1.00 & 1.00 & \textbf{1.00} & 0.67 & 1.00 & 0.80  & 43 \\
 Resource Development & 1.00 & 0.98 & \textbf{0.99} & 0.54 & 0.31 & 0.39 & 45  \\
 
\end{tabular}
\label{table:mitrepertactic}
\end{center}
\end{table*}

We further illustrated the multi-label, pair-wise overlap situation of the predicted results between SecureBERT-SFT and GPT-3.5 in the overlap diagrams shown in Figure \ref{fig:MITREoverlapResults}. Comparing to the actual overlaps (top) given by ATT\&CK, it is evident that GPT-3.5 tends to predict much more loosely and broadly when interpreting the TTP descriptions, while SecureBERT-SFT has a relatively stronger focus to predict individual tactics (within the multi-label classification task). The inherent ambiguity and overlaps of the TTP descriptions exacerbates when GPT-3.5 is used and predicts many overlaps. One may notice that SecureBERT-SFT was not able to predict well for `Lateral Movement' (only 1 prediction for itself and 2 overlaps with `Defense-Evasion' -- very poor recall). This coincides with our conjecture earlier where `Lateral-Movement' descriptions actually read more like other tactics. 


Although SecureBERT-SFT clearly outperformed GPT-3.5 on the ATT\&CK descriptions, we posed the question of whether this advantage would hold when applied to other cybersecurity texts, such as CAPEC descriptions.

\vspace*{-4pt}

\subsection{LLM vs. SFT of small-scale-LLMs (CAPEC)}

To continue our comparison of SFT of small-scale-LLMs and LLMs, we turned our attention to the CAPEC descriptions, assessing the effectiveness of both SecureBERT-SFT and GPT-3.5 models when interpreting indirectly related cyber attack descriptions. The performance results for these models are shown in Tables \ref{table:capecoverall} and \ref{table:capecpertactic}. It can be seen that neither model exhibited the same level of strong performance with these descriptions. A ``0.00'' score for Precision, Recall, and F1 indicates that the models failed to predict the correct tactics. This clearly shows the limitations of the state-of-the-art LLMs when used to interpret the multi-label, ambiguous TTP types of descriptions. We should note that in GPT-3.5 predictions, it labeled one of CAPEC's descriptions as "Social-Engineering". We disregarded this label and replaced it with `NONE' to keep the consistency within the 14 ATT\&CK tactics, and to provide a comparison with SecureBERT-SFT predictions. 

\begin{table}[htbp]
\small
\begin{center}
\caption{Overall performance of SecureBERT-SFT vs. GPT-3.5 for all CAPEC descriptions.}
\begin{tabular}{c | c  c} 
\textbf{Model:} & \textbf{SecureBERT-SFT} & \textbf{GPT-3.5} \\ 
\hline
\textbf{Micro-F1} & \textbf{0.46} & 0.42 \\
\hline
 \textbf{Accuracy} & 0.30 & 0.30 \\
\end{tabular}
\label{table:capecoverall}
\end{center}
\end{table}

A more detailed examination of the results, we see that for the `Execution', `Exfiltration', and `Resource-Development' tactics, even though the number of descriptions were just a few, both models performed very poorly with "0.00" F1 scores. In fact, SecureBERT-SFT overfitted to ATT\&CK as it had 85\% accuracy on ATT\&CK but 30\% on CAPEC descriptions. This is the limitation of SFT of small-scale-LLMs as they rely on patterns in training data to optimize the loss function and struggle to generalize effectively on other descriptions (e.g., CAPEC). 

\begin{figure}
     \centering
     \begin{subfigure}[b]{0.4\textwidth}
         \centering
         \includegraphics[width=\textwidth]{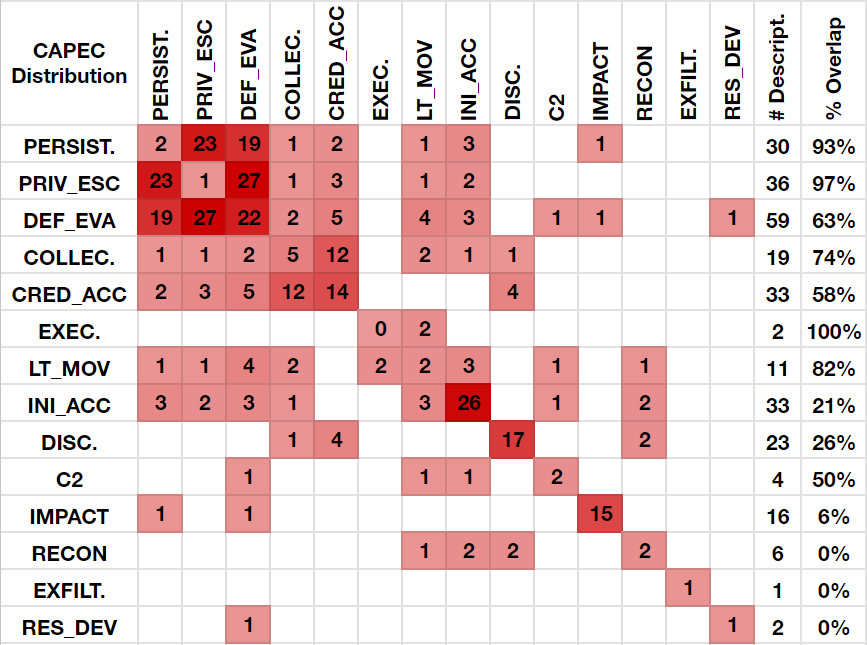}
         \caption{Actual overlaps of CAPEC descriptions}
         \label{fig:originalOverlapCAPEC}
     \end{subfigure}
     \hfill
     \begin{subfigure}[b]{0.4\textwidth}
         \centering
         \includegraphics[width=\textwidth]{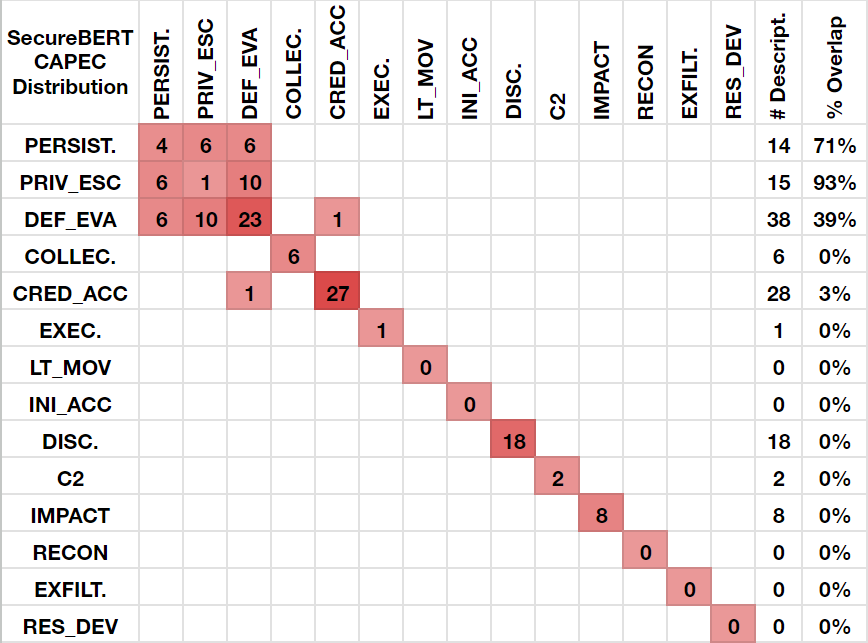}
         \caption{Tactic overlaps based on SecureBERT predictions}
         \label{fig:secureBERToverlapCAPEC}
     \end{subfigure}
     \hfill
     \begin{subfigure}[b]{0.4\textwidth}
         \centering
         \includegraphics[width=\textwidth]{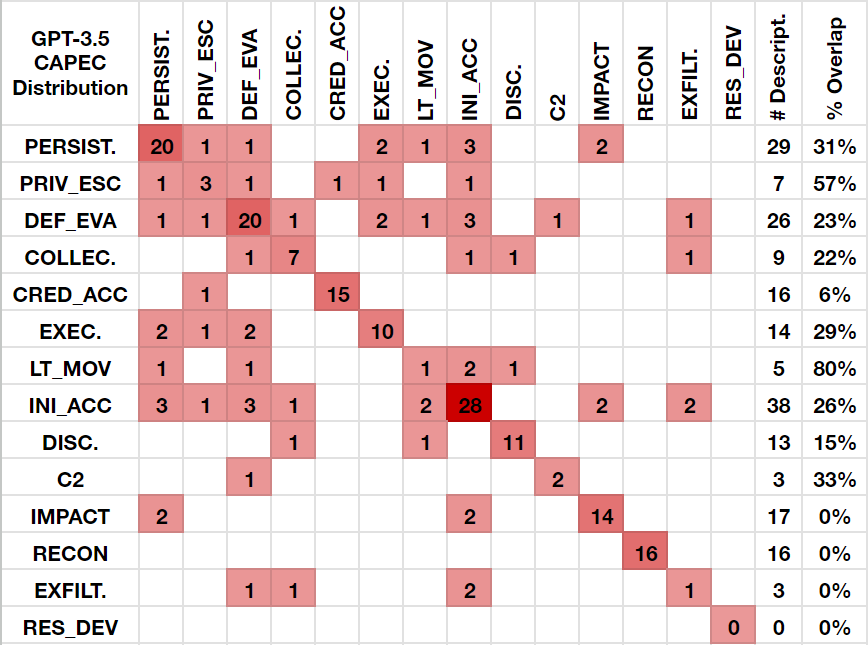}
         \caption{Tactic overlaps based on GPT-3.5 predictions}
         \label{fig:GPT3overlapCAPEC}
     \end{subfigure}
        \caption{Pair-wise tactic overlaps for CAPEC descriptions.}
        \label{fig:CAPECoverlapResults}
\end{figure}

Furthermore, SecureBERT-SFT did not perform well for the `Initial-Access' and `Lateral-Movement' descriptions, either, which have a higher number of descriptions. As shown in the overlaps given by CAPEC in Figure \ref{fig:originalOverlapCAPEC}, there exists a high overlap within the CAPEC descriptions, indicating the even higher level of ambiguity within its descriptions, as both SecureBERT and GPT-3.5 could not perform well on this dataset. However, SecureBERT-SFT has more ``0.00'' scores for CAPEC since it aims at achieving “precise” prediction, while GPT-3.5 tends to produce “casual” overlapping predictions, and therefore, in this case, we argue that GPT-3.5 performs better than SecureBERT-SFT for CAPEC descriptions (despite having an equal accuracy).

\begin{table*}[t]
\small
\begin{center}
\caption{Per-tactic performance of SecureBERT-SFT vs. GPT-3.5 for all CAPEC descriptions.}
\begin{tabular}{@{}c !{\vrule width 1pt} c | c | c !{\vrule width 1pt} c | c | c !{\vrule width 1pt} c@{}} 
  & \multicolumn{3}{c}{\textbf{SecureBERT-SFT}} & \multicolumn{3}{c}{\textbf{GPT-3.5}} \\ 
\hline
\textbf{Tactics} & \textbf{Precision} & \textbf{Recall} & \textbf{F1 Score} & \textbf{Precision} & \textbf{Recall} & \textbf{F1 Score} & \textbf{Support} \\ 
\hline
 Collection & 0.67 & 0.21 & 0.32 & 0.56 & 0.26 & \textbf{0.36} & 19 \\
 Command and Control & 1.00 & 0.50 & \textbf{0.67} & 0.33 & 0.25 & 0.29 & 4  \\
 Credential Access & 0.50 & 0.42 & \textbf{0.46} & 0.69 & 0.33 & 0.45 & 33 \\
 Defense Evasion & 0.66 & 0.42 & \textbf{0.52} & 0.63 & 0.29 & 0.40 & 59 \\
 Discovery & 0.94 & 0.74 & \textbf{0.83} & 0.69 & 0.39 & 0.50 & 23\\
 Execution & 0.00 & 0.00 & 0.00 & 0.00 & 0.00 & 0.00 & 2 \\
 Exfiltration & 0.00 & 0.00 & 0.00 & 0.00 & 0.00 & 0.00 & 1 \\
 Impact & 1.00 & 0.50 & 0.67 & 0.71 & 0.75 & \textbf{0.73} & 16 \\
 Initial Access & 0.00 & 0.00 & 0.00 & 0.53 & 0.61 & \textbf{0.56} & 33  \\
 Lateral Movement & 0.00 & 0.00 & 0.00 & 0.20 & 0.09 & \textbf{0.13} & 11 \\
 Persistence & 0.71 & 0.33 & 0.45 & 0.52 & 0.50 & \textbf{0.51} & 30 \\
 Privilege Escalation & 0.93 & 0.39 & \textbf{0.55} & 0.71 & 0.14 & 0.23 & 36  \\
 Reconnaissance & 0.00 & 0.00 & 0.00 & 0.19 & 0.50 & \textbf{0.27}  & 6 \\
 Resource Development & 0.00 & 0.00 & 0.00 & 0.00 & 0.00 & 0.00 & 2  \\
 
\end{tabular}
\label{table:capecpertactic}
\end{center}
\end{table*}

The bottom two diagrams in Figure \ref{fig:CAPECoverlapResults} show the overlaps based on the predictions of the CAPEC descriptions by both SecureBERT-SFT and GPT-3.5. GPT-3.5 demonstrated a somewhat superior capability in capturing the inherent ambiguity in the descriptions. In contrast, SecureBERT-SFT leaned towards differentiating the descriptions. This observation indicates the different behaviors and strengths of these two LLM approaches in handling complex and ambiguous cybersecurity descriptions.

Table \ref{table:numspecificcases} shows how SecureBERT-SFT and GPT-3.5 were both correct, both wrong, one correct and the other wrong, or partially correct, etc. for the CAPEC descriptions. This gives a perspective where each LLM approach has its advantages and disadvantages. Some of the CAPEC descriptions (50 of them) are interpreted at least partially correct by both, while 61 of them are entirely wrong -- neither model predicts any tactic correctly. These 61 instances (out of 177) presents the absolute limitations of the current LLM capabilities. This results suggest a potential research direction by utilizing an ensemble of the two approaches.

Overall, from all the results presented above, we conclude that when labeled training data is available and representative of the attack description distribution, SFT of small-scale-LLMs is the better approach; otherwise, the use of state-of-the-art LLMs (e.g., GPT-4 or better ones) with proper prompt engineering and adaptation can provide cybersecurity professional a reference (but not as an accurate prediction tool).

\begin{table}[htbp]
\small
\begin{center}
\caption{Specific num. of prediction of SecureBERT-SFT and GPT-3.5 for all CAPEC descriptions.}
\begin{tabular}{c | c} 
\textbf{SecureBERT-SFT (A) \& GPT-3.5 (B)} & \textbf{Count} \\ 
\hline
 A \& B Correct & 29 \\
\hline
 A \& B Wrong & 61  \\
 \hline
 A \& B Partially Correct & 21  \\
 \hline
 A Correct, B Wrong & 15 \\
 \hline
 A Wrong, B Correct &  24 \\
 \hline
 A Partially Correct, B Wrong & 6 \\
 \hline
 A Wrong, B Partially Correct & 13 \\
 \hline
 A Partially Correct, B Correct & 0 \\
 \hline
 A Correct, B Partially Correct & 9 \\
\end{tabular}
\label{table:numspecificcases}
\end{center}
\end{table}

Lastly, we tested the 100 samples of IMDB reviews on both of these models. The GPT-3.5 model correctly classified all of them, and the SecureBERT-SFT model predicted 99 of them as `NONE'. These results provide a high-level sense that both LLM approaches are able to differentiate well clearly non-cybersecurity text with the TTP descriptions. Additional experiments may assess the gray areas between clearly non-cyber text with cybersecurity text describing ATT\&CK tactics. In the following section, we will show some specific descriptions to begin revealing that the gray areas, and thus the limitations of LLMs.

\subsection{Specific Cases}
\label{sec:cases}

As previously discussed, the SecureBERT-SFT provides a more focused and clearer differentiation within descriptions, while the GPT-3.5 tend to produce more overlaps within descriptions. To illustrate this characteristic difference and the inherent ambiguity in the descriptions, we have highlighted some descriptions that show different performance across the two models. 

\vspace{6pt}
\noindent \textbf{Case 1:} The CAPEC description shown below correspond to `Transparent Proxy Abuse' which is mapped to ATT\&CK's `Command-and-Control (C2)' tactic.

\vspace{4pt}
\begin{quote}
\it{A transparent proxy serves as an intermediate between the client and the internet at large. It intercepts all requests originating from the client and forwards them to the correct location. The proxy also intercepts all responses to the client and forwards these to the client. All of this is done in a manner transparent to the client.
}
\end{quote}
\vspace{4pt}

SecureBERT correctly identified the tactic for this case, whereas GPT-3.5 predicted it as `NONE' -- meaning not recognizing it as any ATT\&CK tactic. The description in this particular example is relatively generic, composed of frequently used words (tokens), hence not interpreted by GPT-3.5 as a cyberattack activity. However, SecureBERT-SFT, having been explicitly trained for this task, pays more attention to specific tokens in the text, such as `proxy', which is the name for one of the techniques of C2 in the ATT\&CK framework. This can justify the correct prediction of SecureBERT-SFT for this specific example.

\vspace{6pt}
\noindent \textbf{Case 2:} The ATT\&CK description shown below correspond to ATT\&CK's `Lateral-Movement' tactic.
\begin{quote}
\it{Adversaries may exploit remote services to gain unauthorized access to internal systems once inside of a network. Exploitation of a software vulnerability occurs when an adversary takes advantage of a programming error in a program, service, or within the operating system software or kernel itself to execute adversary-controlled code.  A common goal for post-compromise exploitation of remote services is for lateral movement to enable access to a remote system. An adversary may need to determine if the remote system is in a vulnerable state, which may be done through Network Service Discovery or other Discovery methods looking for common, vulnerable software that may be deployed in the network, the lack of certain patches that may indicate vulnerabilities, or security software that may be used to detect or contain remote exploitation. Servers are likely a high value target for lateral movement exploitation, but endpoint systems may also be at risk if they provide an advantage or access to additional resources. There are several well-known vulnerabilities that exist in common services such as SMB and RDP as well as applications that may be used within internal networks such as MySQL and web server services. Depending on the permissions level of the vulnerable remote service an adversary may achieve Exploitation for Privilege Escalation as a result of lateral movement exploitation as well. }
\end{quote}

In this case, SecureBERT was unable to recognize it as an MITRE ATT\&CK tactic, classifying it as `NONE'. on the other hand, GPT-3.5, predicted `Lateral-Movement', `Initial-Access', `Discovery', and `Privilege-Escalation' as the predicted tactics. This long and ambiguous description covers quite a few concepts for an ordinary human reader. Therefore, SecureBERT failed to confidently determine keywords (tokens) with high attention values for specific tactics, while GPT-3.5 interprets it broadly and covers four tactics. However, in this case, GPT-3.5 performed better as it correctly identified `Lateral-Movement' (along with other tactics), which is better than not identifying any.

\vspace{6pt}
\noindent \textbf{Case 3:} The CAPEC description shown below correspond to `Inducing-Account-Lockout' -- mapped to ATT\&CK's `Impact' tactic.

\begin{quote}
\it{An attacker leverages the security functionality of the system aimed at thwarting potential attacks to launch a denial of service attack against a legitimate system user. Many systems, for instance, implement a password throttling mechanism that locks an account after a certain number of incorrect log in attempts. An attacker can leverage this throttling mechanism to lock a legitimate user out of their own account. The weakness that is being leveraged by an attacker is the very security feature that has been put in place to counteract attacks.
}
\end{quote}
\vspace{1pt}

For this description, SecureBERT predicted it as `NONE', whereas GPT-3.5 predicted it as `Defense-Evasion' -- both failed to predict `Impact', the human expert assigned ATT\&CK tactic. One may argue that this is a limitation for both of these models, whereas the description simply talks about an attacker technique to cause user not able to access their account. While the description is indeed related to `Impact', no training data from ATT\&CK contains related keywords (tokens) and GPT-3.5 clearly missed the linkage between the technique itself and the effect of the technique. This shows the need for a more powerful and specialized cybersecurity LLM to comprehend the ambiguity in descriptions better.

The examples provided above clearly underscore the presence of ambiguity in cybersecurity descriptions. This ambiguity often confounds both the SFT of small-scale-LLMs and the LLMs, making it challenging for them to accurately interpret and describe the content.  In instances where the explanations possess inherent generality (as in Case 2) or requires implicit linkage between techniques to consequences (Case 3), neither the SecureBERT-SFT nor GPT-3.5 is able to interpret the descriptions. 
\noindent This leaves us to ponder an important question: How can we effectively address and deal with this inherent ambiguity? We will talk about the current limitations and our recommendations to address these challenges in the next section. 



\section{Discussion: from Challenges to Recommendations}
\label{sec:recomm}

\subsection{Challenges: LLMs for TTP interpretation}
\label{sec:challenges}


The application of LLMs to the interpretation of TTPs, poses several challenges. Inherent ambiguities in TTP descriptions are often introduced due to human interpretation variances when TTPs are documented and reported by professionals and journalists. These ambiguities can be classified into different types, each requiring its own unique treatment. Type 1 Ambiguity is caused by the implication of numerous tactics in the same description. A single phrase or sentence may suggest multiple possible tactics, making it challenging for LLMs to decipher the intended course of action. Type 2 Ambiguity can arise when a description contains both the approach (i.e., how) and the intended effect (i.e., why). Therefore, it becomes challenging for LLMs to differentiate between the method used by the attacker and the purpose behind it. Another point of contention lies in the understanding of ``Impact''. The term often signifies the effects on a user or an organization, making it a unique concept that differs substantially from the effect of a technique or sub-technique. Therefore, differentiating between ``Impact'' and other `tactics' effects poses an additional layer of complexity for LLMs. These challenges underscore the need for continuous advancements and refinements in the application of LLMs to the interpretation of TTPs. 


Meanwhile, as an unintended consequence, this study also reveals the needs to improve the clarity of MITRE ATT\&CK framework. Considering the various cybersecurity enumerations and taxonomies proposed over the years, MITRE ATT\&CK is arguably most widely adopted in various SIEM tools. Over the past few years, many techniques, sub-techniques, and procedures were added, with each mapping to, more often than not, multiple tactics. Understandably, this is due to the many possibilities of how an adversary can use an attack technique. Unfortunately, this also leads to the aforementioned ambiguity, for both human and LLMs. It is worth questioning how the community should proceed in maintaining and updating the framework. Continuing with the 14 tactics? Adding new techniques and procedures with caution? It will be imperatives NOT to render an overly complex and ambiguous ATT\&CK unusable in the near future.



Regardless of how ATT\&CK will continue to evolve, we expect there will be inherent ambiguity in TTP descriptions, and the community still needs directions for the proper uses of LLMs to assist in cybersecurity operations. As discussed in Section 4, SFT of small-scale-LLMs and LLMs approach the task of interpreting the cybersecurity descriptions differently. SFT of small-scale-LLMs aim to distinguish between different tactics, whereas GPT-3.5 is not inherently designed to do so. This difference in the two approaches underlies the performance disparities we have observed, and exacerbates the way they treat the ambiguities. With such, we make the following recommendations for the research community to further investigate and develop optimal uses of LLMs to treat the inherently ambiguous TTP descriptions.

\vspace*{-4pt}
\subsection{Recommendation: Self-Supervised LLMs}
\label{sec:recomm1}

Direct use of GPT models allows us to access the pre-trained LLMs and use their capabilities for various tasks. However, crafting an effective prompt is an important task. Prompt Engineering is an iterative process which involves understanding the context, the right example for the model, and how well to structure it with alleviated ambiguity. Moreover, the capabilities of LLMs extend beyond prompt design. Fine-tuning these models can yield superior results as it allows for training on a greater number of examples than what could typically fit in a prompt \cite{Fine-tuning}. 

Therefore, one direction for improvement could be the fine-tuning of LLMs specifically for cybersecurity applications. Such focused training could potentially enhance the model's ability to differentiate between tactics and reduce ambiguity. Therefore, having a specialized cybersecurity LLM could enhance the ambiguity within cybersecurity descriptions, and could further be applied to different cybersecurity tasks, such as threat analysis and detection. Note that how to effectively fine-tuning specialized LLMs is still an open problem since the scale of the model requires a significant amount of high-quality data. As evidenced by the underwhelming performance of SecureBERT, it is insufficient by fine-tuning small-scale LLMs.

Moreover, the inherent ambiguity in LLMs may decrease through approaches like Reinforcement Learning with Human Feedback \cite{christiano2017deep}. However, this ambiguity can manifest both in the prompts given to the model and in its responses, leading to potential misinterpretations. There is also the risk of adversarial attacks exploiting this ambiguity. We can study examples of how ambiguity can be weaponized to mislead or break the system. Documenting such instances is crucial for understanding the model's vulnerabilities and designing protective measures.

Due to the continuous evolving nature of vulnerabilities and exploits, there is a need to process new/unseen vulnerabilities and represent them in a useful way to gain an understanding of the tactics and techniques utilized by adversaries. Therefore, Continual Learning could be a potential avenue for future exploration. By continually updating and learning from new cybersecurity incidents, the model can stay current and improve its performance. Therefore, with open-source LLMs being publicly available, such as LLAMA-2 \cite{touvron2023llama}, allows for continuous fine-tuning of these LLMs for TTP interpretation. Note that SFT of LLMs like LLAMA-2 requires a huge amount of computational resources. Therefore, to deal with this problem, there have been parameter efficient fine-tuning (PEFT) algorithms proposed that makes the fine-tuning of this LLMs possible with less training parameters, such as LoRA \cite{hu2021lora} and QLoRA \cite{dettmers2023qlora}.

Furthermore, the use of self-supervised metrics to detect adversarial content and providing explanatory comments can be another future exploration. Finally, while models like LLMs can be powerful and accurate, explainability remains a key requirement, especially in a field as critical as cybersecurity. Security analysts must be able to understand and trust the model's predictions. The need for explainability of LLM recommendations might necessitate the development of additional tools or features, such as text highlighting or pop-up explanations for chosen labels.

\vspace*{-4pt}
\section{Ethical Concerns}
\label{sec:ethics}
We do not believe that there is any ethical concern for this research, since this work does not directly discover new vulnerabilities or exploits. This work focuses on analyzing two practical uses of LLMs to interpret unstructured TTP descriptions. Our findings offers insights on what reasonable expectations cybersecurity professionals should have when using LLMs, and the challenges of the inherent ambiguity in TTP descriptions (ATT\&CK and CAPEC). While there is always possibilities adversaries can take advantage of these ambiguities, it is more important for the community to recognize the limitations and begin to alleviate them.

\vspace*{-4pt}
\section{Concluding Remarks}
\label{sec:conclusion}

This work presented, to the best of our knowledge, the first comparative analysis on practical uses of LLMs to interpret TTP descriptions. We reviewed previous works on LLMs for different cybersecurity applications, and specifically for interpreting cybersecurity texts. We proposed and presented a comprehensive analysis between the direct use of LLMs, GPT-3.5 and Bard, and SFT of small-scale-LLMs, BERT and SecureBERT - both taking into consideration on the real-world constraints cybersecurity practitioners may face when using LLMs. Our study yields insightful observations and notable recommendations for future research in this field.

Our findings revealed much ambiguity and overlaps exist in TTP descriptions. Even as widely-vetted as MITRE ATT\&CK and CAPEC frameworks are, average cyber-op personnel will have a hard time consistently interpreting and communicating the intended tactics/effects of cyberattacks. The rise of LLMs can easily lead to their mis-uses for cyber-ops. Our findings indicated that the SFT of small-scale-LLMs provide a more focused and clearer classification of cybersecurity tactics. Particularly, SecureBERT-SFT outperformed GPT-3.5 and displayed a promising level of accuracy and F1-score in classifying ATT\&CK descriptions. However, when tested with CAPEC descriptions (similar but different than ATT\&CK), we observed that neither is capable of performing at the same level as for the ATT\&CK cases. We observe that the direct use of LLMs, while not accurately predicting ATT\&CK tactics, offers a broader and flexible interpretation of TTP descriptions. Their ability to generate language and comprehend a wider range of Internet texts makes them versatile in interpreting the specialized descriptions, and exhibits an intriguing capability to expose the inherent ambiguity and overlaps within the TTP descriptions. Overall, when labeled training data is available and representative of the attack description distribution, SFT of small-scale-LLMs is the better approach; otherwise, the use of state-of-the-art LLMs (e.g., GPT-4 or better ones) with proper prompt engineering and adaptation can provide cybersecurity professional a reference for further examination.

Our findings also shed light on several potential areas for improvement and further exploration. The need for continual learning of models, the importance of explainability, and the handling of inherent ambiguity all emerge as important considerations for the future of AI/ML in cybersecurity. We provided numerous avenues for improvement in the field of LLM applying to cybersecurity. By addressing these limitations and implementing the recommended adjustments, the community can create more accurate, reliable, and understandable LLM tools for cybersecurity analysis.

In conclusion, our study underscores the complexity of interpreting unstructured, ambiguous cybersecurity descriptions and provides a comparative analysis of two practical uses of LLMs to interpret them. Our observations and recommendations raise the need for more nuanced, effective, and context-aware applications of generative AI in cybersecurity.

\bibliographystyle{acm}
\bibliography{references}

\begin{thebibliography}{10}

\bibitem{AbdulNabi2021}
{\sc AbdulNabi, I., and Yaseen, Q.}
\newblock Spam email detection using deep learning techniques.
\newblock {\em Procedia Computer Science 184\/} (1 2021), 853--858.

\bibitem{Aghaei2023}
{\sc Aghaei, E., Niu, X., Shadid, W., and Al-Shaer, E.}
\newblock {SecureBERT}: A domain-specific language model for cybersecurity.
\newblock {\em Lecture Notes of the Institute for Computer Sciences,
  Social-Informatics and Telecommunications Engineering, LNICST 462 LNICST\/}
  (2023), 39--56.

\bibitem{Alhogail2021}
{\sc Alhogail, A., and Alsabih, A.}
\newblock Applying machine learning and natural language processing to detect
  phishing email.
\newblock {\em Computers \& Security 110\/} (2021), 102414.

\bibitem{Alves2022}
{\sc Alves, P.~M., Geraldo~Filho, P., and Gon{\c{c}}alves, V.~P.}
\newblock Leveraging {BERT}'s power to classify {TTP} from unstructured text.
\newblock {\em IEEE\/} (2022), 1--7.

\bibitem{christiano2017deep}
{\sc Christiano, P.~F., Leike, J., Brown, T., Martic, M., Legg, S., and Amodei,
  D.}
\newblock Deep {Reinforcement} {Learning} from human preferences.
\newblock {\em Advances in neural information processing systems 30\/} (2017).

\bibitem{Costa2022}
{\sc Costa, J.~C., Roxo, T., Sequeiros, J.~B., Proenca, H., and Inacio, P.~R.}
\newblock Predicting cvss metric via description interpretation.
\newblock {\em IEEE Access 10\/} (2022), 59125--59134.

\bibitem{Das2020}
{\sc Das, S., Ashrafuzzaman, M., Sheldon, F.~T., and Shiva, S.}
\newblock Network intrusion detection using natural language processing and
  ensemble machine learning.
\newblock {\em IEEE Symposium Series on Computational Intelligence (SSCI)\/}
  (2020), 829--835.

\bibitem{Oliveira2021}
{\sc de~Oliveira, N.~R., Pisa, P.~S., Lopez, M.~A., de~Medeiros, D. S.~V., and
  Mattos, D.~M.}
\newblock Identifying fake news on social networks based on natural language
  processing: trends and challenges.
\newblock {\em Information 12}, 1 (2021), 38.

\bibitem{dettmers2023qlora}
{\sc Dettmers, T., Pagnoni, A., Holtzman, A., and Zettlemoyer, L.}
\newblock Qlora: Efficient finetuning of quantized {LLMs}.
\newblock {\em arXiv preprint arXiv:2305.14314\/} (2023).

\bibitem{Devlin2018}
{\sc Devlin, J., Chang, M.-W., Lee, K., and Toutanova, K.}
\newblock {BERT}: Pre-training of deep bidirectional transformers for language
  understanding.
\newblock {\em Proceedings of NAACL-HLT\/} (2019), 4171--4186.

\bibitem{Bard-API}
{\sc dsdanielpark/Bard API, G.}
\newblock Bard-api.
\newblock {\em \url{https://github.com/dsdanielpark/Bard-API}\/}.
\newblock Accessed: 2023-05.

\bibitem{ParamsBardGPT}
{\sc Gooding, M.}
\newblock Bard: Google reveals generative ai chatbot to take on chatgpt.
\newblock {\em
  \url{https://techmonitor.ai/technology/ai-and-automation/bard-google-chatgpt-generative-ai-chatbot}\/}
  (2023).
\newblock Accessed: 2023-05.

\bibitem{Bard}
{\sc Google}.
\newblock Bard.
\newblock {\em \url{https://bard.google.com/}\/} (2023).
\newblock Accessed: 2023-04.

\bibitem{Haque2023}
{\sc Haque, M.~A., Shetty, S., Kamhoua, C.~A., and Gold, K.}
\newblock Adversarial technique validation \& defense selection using attack
  graph \& {ATT\&CK} matrix.
\newblock {\em International Conference on Computing, Networking and
  Communications (ICNC)\/} (2023), 181--187.

\bibitem{Heidari2021}
{\sc Heidari, M., Zad, S., Hajibabaee, P., Malekzadeh, M., HekmatiAthar, S.,
  Uzuner, O., and Jones, J.~H.}
\newblock {BERT} model for fake news detection based on social bot activities
  in the {COVID}-19 pandemic.
\newblock {\em IEEE 12th Annual Ubiquitous Computing, Electronics \& Mobile
  Communication Conference (UEMCON)\/} (2021), 0103--0109.

\bibitem{hu2021lora}
{\sc Hu, E.~J., Wallis, P., Allen-Zhu, Z., Li, Y., Wang, S., Wang, L., Chen,
  W., et~al.}
\newblock {LoRA}: Low-rank adaptation of large language models.

\bibitem{Husari2017}
{\sc Husari, G., Al-Shaer, E., Ahmed, M., Chu, B., and Niu, X.}
\newblock {TTPDrill}: Automatic and accurate extraction of threat actions from
  unstructured text of {CTI} sources.
\newblock {\em Proceedings of the 33rd annual computer security applications
  conference\/} (2017), 103--115.

\bibitem{Kaliyar2021}
{\sc Kaliyar, R.~K., Goswami, A., and Narang, P.}
\newblock {FakeBERT}: Fake news detection in social media with a {BERT}-based
  deep learning approach.
\newblock {\em Multimedia Tools and Applications 80\/} (2021), 11765--11788.

\bibitem{Kim2022}
{\sc Kim, H., Kim, H., et~al.}
\newblock Comparative experiment on {TTP} classification with class imbalance
  using oversampling from {CTI} dataset.
\newblock {\em Security and Communication Networks 2022\/} (2022).

\bibitem{Legoy2019}
{\sc Legoy, V. S.~M.}
\newblock Retrieving {ATT\&CK} tactics and techniques in cyber threat reports.

\bibitem{Liu2019}
{\sc Liu, Y., Ott, M., Goyal, N., Du, J., Joshi, M., Chen, D., Levy, O., Lewis,
  M., Zettlemoyer, L., and Stoyanov, V.}
\newblock {RoBERTa}: A robustly optimized {BERT} pre-training approach.
\newblock {\em arXiv preprint arXiv:1907.11692\/} (2019).

\bibitem{CAPEC}
{\sc MITRE}.
\newblock {CAPEC} - common attack pattern enumeration and classification.
\newblock {\em \url{https://capec.mitre.org/index.html}\/}.
\newblock Accessed: 2023-03.

\bibitem{CVE}
{\sc MITRE}.
\newblock {CVE} - common vulnerabilities and exposures.
\newblock {\em \url{https://cve.mitre.org/}\/}.
\newblock Accessed: 2023-02.

\bibitem{MITRE}
{\sc MITRE}.
\newblock {MITRE ATT\&CK®}.
\newblock {\em \url{https://attack.mitre.org/}\/}.
\newblock Accessed: 2023-02.

\bibitem{IMDB}
{\sc N, L.}
\newblock {IMDB} dataset of 50k movie reviews.
\newblock {\em
  \url{https://www.kaggle.com/datasets/lakshmi25npathi/imdb-dataset-of-50k-movie-reviews}\/}.
\newblock Accessed: 2023-04.

\bibitem{Fine-tuning}
{\sc {OpenAI}}.
\newblock {Fine-Tuning} - {OpenAI API}.
\newblock {\em \url{https://platform.openai.com/docs/guides/fine-tuning}\/}.
\newblock Accessed: 2023-05.

\bibitem{OpenAI}
{\sc {OpenAI}}.
\newblock Models - {OpenAI API}.
\newblock {\em \url{https://platform.openai.com/docs/models/gpt-3-5}\/}.
\newblock Accessed: 2023-04.

\bibitem{GPT3}
{\sc {OpenAI}}.
\newblock {OpenAI's GPT3}.
\newblock {\em \url{https://openai.com/blog/gpt-3-apps}\/}.
\newblock Accessed: 2023-02.

\bibitem{Orbinato2022}
{\sc Orbinato, V., Barbaraci, M., Natella, R., and Cotroneo, D.}
\newblock Automatic mapping of unstructured cyber threat intelligence: An
  experimental study:(practical experience report).
\newblock {\em IEEE 33rd International Symposium on Software Reliability
  Engineering (ISSRE)\/} (2022), 181--192.

\bibitem{Oshikawa2020}
{\sc Oshikawa, R., Qian, J., and Wang, W.~Y.}
\newblock A survey on natural language processing for fake news detection.
\newblock {\em Proceedings of the 12th Language Resources and Evaluation
  Conference\/} (2020), 6086--6093.

\bibitem{Rahali2021}
{\sc Rahali, A., and Akhloufi, M.~A.}
\newblock {MalBERT}: Using transformers for cybersecurity and malicious
  software detection.
\newblock {\em arXiv preprint arXiv:2103.03806\/} (2021).

\bibitem{Rahali2023}
{\sc Rahali, A., and Akhloufi, M.~A.}
\newblock {MalBERTv2}: Code aware {BERT}-based model for malware
  identification.
\newblock {\em Big Data and Cognitive Computing 7}, 2 (2023), 60.

\bibitem{Rahman2022}
{\sc Rahman, M.~R., and Williams, L.}
\newblock Investigating co-occurrences of {MITRE ATT\&CK} techniques.
\newblock {\em arXiv preprint arXiv:2211.06495\/} (2022).

\bibitem{Rani2023}
{\sc Rani, N., Saha, B., Maurya, V., and Shukla, S.~K.}
\newblock {TTPHunter}: Automated extraction of actionable intelligence as
  {TTP}s from narrative threat reports.
\newblock {\em Proceedings of the 2023 Australasian Computer Science Week\/}
  (2023), 126--134.

\bibitem{reimers2019sentence}
{\sc Reimers, N., and Gurevych, I.}
\newblock Sentence-{BERT}: Sentence embeddings using siamese {BERT}-networks.
\newblock {\em arXiv preprint arXiv:1908.10084\/} (2019).

\bibitem{Sauerwein2022}
{\sc Sauerwein, C., and Pfohl, A.}
\newblock Towards automated classification of attackers' {TTP}s by combining
  {NLP} with {ML} techniques.
\newblock {\em arXiv preprint arXiv:2207.08478\/} (2022).

\bibitem{Shahid2021}
{\sc Shahid, M.~R., and Debar, H.}
\newblock {CVSS-BERT}: Explainable natural language processing to determine the
  severity of a computer security vulnerability from its description.
\newblock {\em 20th IEEE International Conference on Machine Learning and
  Applications (ICMLA)\/} (2021), 1600--1607.

\bibitem{Suciu2022}
{\sc Suciu, O., Nelson, C., Lyu, Z., Bao, T., and Dumitraș, T.}
\newblock Expected exploitability: Predicting the development of functional
  vulnerability exploits.
\newblock {\em 31st USENIX Security Symposium (USENIX Security 22)\/} (2022),
  377--394.

\bibitem{sun2019chinese}
{\sc Sun, C., Qiu, X., Xu, Y., and Huang, X.}
\newblock How to fine-tune {BERT} for text classification?
\newblock {\em Chinese Computational Linguistics: 18th China National
  Conference, CCL 2019, Kunming, China, October 18--20, 2019, Proceedings 18\/}
  (2019), 194--206.

\bibitem{Tamkin2021}
{\sc Tamkin, A., Brundage, M., Clark, J., and Ganguli, D.}
\newblock Understanding the capabilities, limitations, and societal impact of
  large language models.
\newblock {\em arXiv preprint arXiv:2102.02503\/} (2021).

\bibitem{Teubner2023}
{\sc Teubner, T., Flath, C.~M., Weinhardt, C., van~der Aalst, W., and Hinz, O.}
\newblock Welcome to the era of {ChatGPT} et al. the prospects of large
  language models.
\newblock {\em Business \& Information Systems Engineering 65}, 2 (4 2023),
  95--101.

\bibitem{touvron2023llama}
{\sc Touvron, H., Martin, L., Stone, K., Albert, P., Almahairi, A., Babaei, Y.,
  Bashlykov, N., Batra, S., Bhargava, P., Bhosale, S., et~al.}
\newblock Llama 2: Open foundation and fine-tuned chat models.
\newblock {\em arXiv preprint arXiv:2307.09288\/} (2023).

\bibitem{Yesir2021}
{\sc Yesir, S., and So{\u{g}}ukpinar, {\.I}.}
\newblock Malware detection and classification using fasttext and {BERT}.
\newblock 1--6.

\bibitem{Yin2020}
{\sc Yin, J., Tang, M.~J., Cao, J., and Wang, H.}
\newblock Apply transfer learning to cybersecurity: Predicting exploitability
  of vulnerabilities by description.
\newblock {\em Knowledge-Based Systems 210\/} (12 2020), 106529.

\bibitem{TRAM}
{\sc Yoder, S.}
\newblock Automating mapping to {ATT\&CK}: The threat report {ATT\&CK} mapper
  ({TRAM}) tool.
\newblock {\em
  \url{https://medium.com/mitre-attack/automating-mapping-to-attack-tram-1bb1b44bda76}\/}.
\newblock Accessed: 2023-02.

\bibitem{You2022}
{\sc You, Y., Jiang, J., Jiang, Z., Yang, P., Liu, B., Feng, H., Wang, X., and
  Li, N.}
\newblock {TIM}: threat context-enhanced {TTP} intelligence mining on
  unstructured threat data.
\newblock {\em Cybersecurity 5}, 1 (2022), 3.

\bibitem{Zhao2023}
{\sc Zhao, W.~X., Zhou, K., Li, J., Tang, T., Wang, X., Hou, Y., Min, Y.,
  Zhang, B., Zhang, J., Dong, Z., et~al.}
\newblock A survey of large language models.
\newblock {\em arXiv preprint arXiv:2303.18223\/} (3 2023).

\bibitem{zhuang2022understanding}
{\sc Zhuang, Z., Liu, M., Cutkosky, A., and Orabona, F.}
\newblock Understanding adamw through proximal methods and scale-freeness.
\newblock {\em arXiv preprint arXiv:2202.00089\/} (2022).

\end{thebibliography}



\end{document}